\documentclass[preprint,12pt]{elsarticle}

\usepackage{amssymb}
\usepackage{amsmath}
\usepackage{tabularx} 
\usepackage{makecell} 
\usepackage{graphicx}
\usepackage{float}
\usepackage{multirow} 
\usepackage{booktabs} 

\journal{arXiv}

\begin{document}

\begin{frontmatter}



\title{SleepGMUformer: A gated multimodal temporal neural network for sleep staging}


\author[1]{Chenjun Zhao}
\ead{zhaochenjun@qdu.edu.cn}
\author[1]{Xuesen Niu}
\ead{niuxuesen@qdu.edu.cn}
\author[1]{Xinglin Yu}
\ead{yuxinglin@qdu.edu.cn}
\author[2]{Long Chen}
\ead{lchen6@ic.ac.uk}
\author[3]{Na Lv}
\ead{ise lvn@ujn.edu.cn}
\author[4]{Huiyu Zhou}
\ead{hz143@leicester.ac.uk}
\author[1]{Aite Zhao \corref{cor1}}
\ead{zhaoaite@qdu.edu.cn}
\cortext[cor1]{Corresponding author}
\address[1]{College of Computer Science and Technology, Qingdao University, Qingdao, China}
\address[2]{Institute of Clinical Sciences, Faculty of Medicine, Imperial College London, U.K.}
\address[3]{School of Information Science and Engineering, University of Jinan, Jinan, China}
\address[4]{School of Computing and Mathematical Sciences, University of Leicester, U.K.}

\begin{abstract}
Sleep staging is a key method for assessing sleep quality and diagnosing sleep disorders. However, current deep learning methods face challenges: 1) post-fusion techniques ignore the varying contributions of different modalities; 2) unprocessed sleep data can interfere with frequency-domain information.
To tackle these issues, this paper proposes a gated multimodal temporal neural network for multidomain sleep data, including heart rate, motion, steps, EEG (Fpz-Cz, Pz-Oz), and EOG from WristHR-Motion-Sleep and SleepEDF-78. The model integrates: 1) a pre-processing module for feature alignment, missing value handling, and EEG de-trending; 2) a feature extraction module for complex sleep features in the time dimension; and 3) a dynamic fusion module for real-time modality weighting.Experiments show classification accuracies of 85.03\% on SleepEDF-78 and 94.54\% on WristHR-Motion-Sleep datasets. The model handles heterogeneous datasets and outperforms state-of-the-art models by 1.00\%-4.00\%.

\end{abstract}


\end{frontmatter}


\section{Introduction}
\label{sec:introduction}
Sleep is a pivotal procedure for human health \cite{ref1} and is associated with mood, behavior, and physiological functions such as memory and cognition \cite{ref2}. Sleep staging is a central aspect of sleep assessment and research the accuracy of sleep staging is not only relevant to the assessment of sleep quality \cite{ref3} but also key to achieving early intervention for sleep disorders and related psychiatric disorders \cite{ref4}.

Polysomnography is a multi-parameter study of sleep \cite{ref5}, a test to diagnose sleep disorders through different types of physiological signals recorded during sleep, such as electroencephalography (EEG), cardiography (CG), electrooculography (EOG), electromyography (EMG), oro-nasal airflow and oxygen saturation \cite{ref6}. According to the Rechtschaffen and Kales (R\&K) rule, PSG signals are usually divided into 30-second segments and classified into six sleep stages, namely wakefulness (Wake), four non-rapid eye movement stages (i.e., S1, S2, S3, and S4), and rapid eye movement (REM). In 2007, the American Academy of Sleep Medicine (AASM) adopted the Rechtschaffen \& Kales (R\&K) sleep staging system for Non-Rapid Eye Movement (NREM) sleep. 
\begin{figure}[t]
\centering

\includegraphics[width=0.9\textwidth]{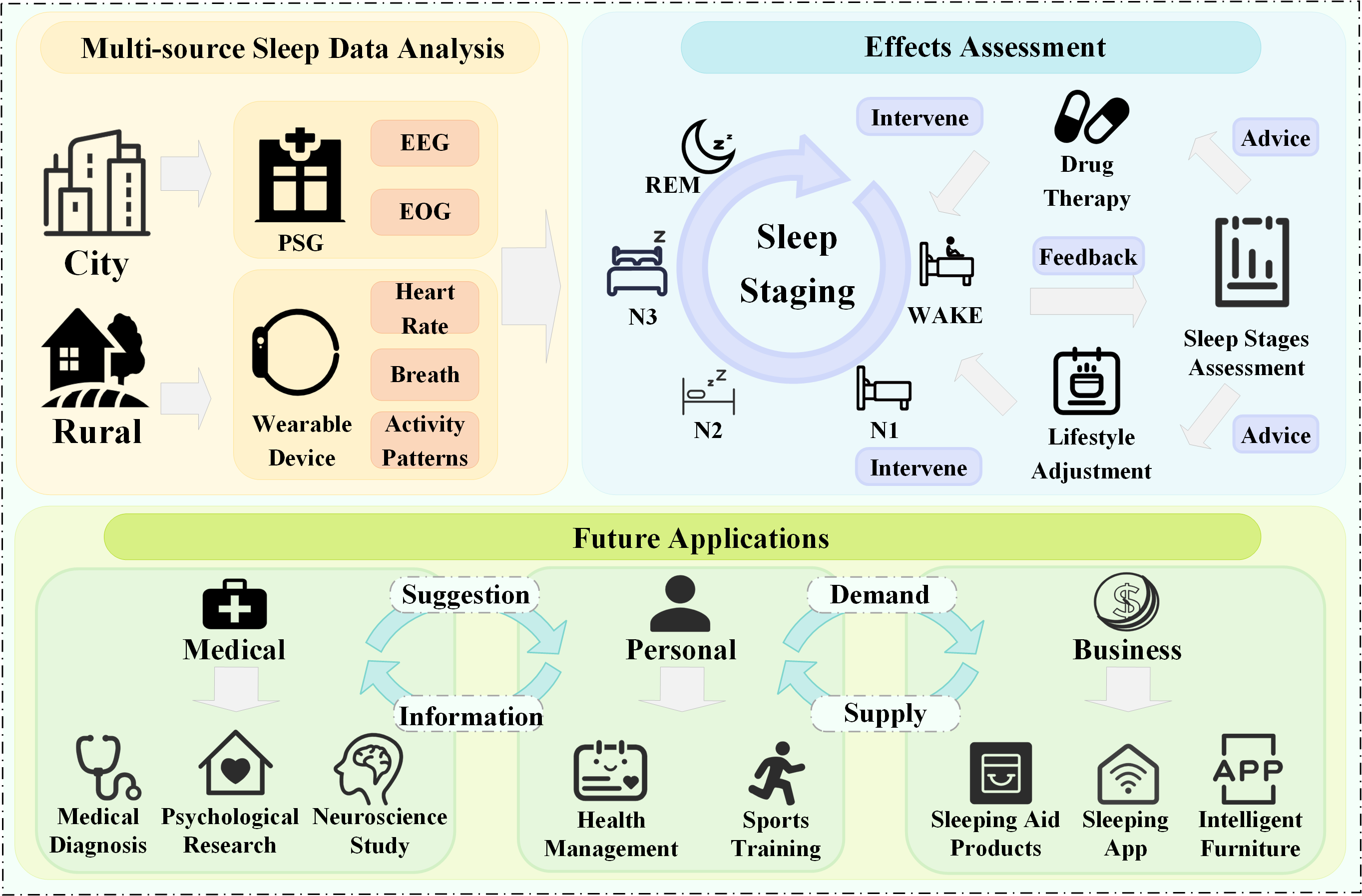}
\caption{Flowchart of sleep staging by integrating SleepGMUformer, including completion of urban and rural sleep data collection and analysis through the use of different devices and data-compatible processing with SleepGMUformer;  Cyclic dynamic assessment and analysis of physical therapy effects is completed through the interaction of drug therapy, lifestyle adjustment and sleep staging assessment; Finally, a vision of future applications involving the medical field, the personal care field, and the commercial field.}\label{fig1}

\end{figure}
Sleep specialists typically utilize these criteria for the manual classification of sleep stages, a process that is not only labor-intensive but also prone to subjective bias \cite{ref7}. Therefore, automated sleep staging is a more efficient alternative to manual methods and has more clinical value \cite{ref8}. Meanwhile, due to the shortcomings of polysomnography itself, on the one hand, the cost and resource consumption will make it prohibitive to implement polysomnography in areas with backward medical conditions, on the other hand, the need to use multiple electrodes and sensors during the test will cause physical discomfort and affect the patient's sleep quality. Therefore, people have begun to study the use of wearable devices for sleep detection, through the Apple Watch to collect relevant sleep data.

Simple deep learning networks are unable to capture time-varying features and time-series information in sleep signals.To overcome these limitations, Supratak et al. proposed DeepSleepNet, an innovative deep-learning model that combines the strengths of CNNs and Bi-LSTMs \cite{ref13}. DeepSleepNet is capable of simultaneously extracting both time-invariant features and learning time-series information for a more comprehensive understanding and analysis of sleep data. Although DeepSleepNet has made breakthroughs in the field of sleep staging, it mainly adopts a one-to-one input-output model, which is direct and transparent but ignores the transition rules that exist between different sleep stages. To solve this problem, Phan et al. proposed SeqSleepNet, a many-to-many training model that accepts multiple sleep stage sequences as inputs and learns the transition rules between sleep stages considering the contextual relationships. The proposal of SeqSleepNet addresses the shortcomings of DeepSleepNet in dealing with sequential data and provides a richer set of transition rules for sleep staging provides richer time series analysis \cite{ref14}. However, SeqSleepNet mainly employs the LSTM network, and although LSTM is intuitive for processing PSG signal sequences, it does not take into account the distribution of EEG electrodes in non-Euclidean space, thus ignoring the spatial correlation between electrodes. 

To address this challenge, Li et al. proposed a combination of dynamic and static spatiotemporal map convolutional networks combining multiple temporal attention blocks \cite{attention}. This approach effectively captured the long-term dependencies between different EEG signals, providing superior performance for sleep staging and compensating for the lack of spatial information processing in SeqSleepNet \cite{ref15}. Despite the remarkable success of SeqSleepNet in the field of sleep staging, it still focuses on single-modality data and cannot effectively integrate and utilize multimodal information \cite{ref16}.

Relying on a single fixed physiological signal is not the best way to differentiate between specific sleep stages due to the limited performance demonstrated by methods based on single-channel EEG signals. In contrast, PSG has a multichannel structure that improves the accuracy, as well as the interpretability of the model's results for classifying sleep stages, and multimodal approaches, are beginning to be adopted \cite{ref17}. Multimodal learning is a method of learning using data from a variety of different sensors or interaction modes. The key to multimodal learning is to integrate and analyze data from different sources to gain more comprehensive and deeper insights than a single data source. Based on a multi-channel scheme, Dong et al. combined DNN and RNN to extract salient features from EEG and EOG signals \cite{ref18}.Chambon et al. proposed a neural network using multi-channel multimodal signals (EEG, EOG, and EMG) as input data. These methods are mainly derived from multichannel features in PSG signals and are combined by means of concatenation. However, the way of concatenating channel signal features ignores the differences in the contribution of multiple modalities to sleep stages, which affects the overall performance of the sleep model \cite{ref19}. Recent studies have also fully integrated multimodal information and demonstrated the different contributions of each modality in identifying specific sleep stages.

In addition, most of the methods are only designed specifically for PSG data that suffers from the shortcomings of exorbitance, the need for monitoring to be carried out in a specific sleep center, and the need for multiple electrodes and sensors to be installed on the patient during the monitoring process \cite{ref22}. Wearable sensors, however, have the advantages of portability, negligible interference with patient repose, and cost-effective \cite{ref23}. Sleep data (including heart rate and respiration) obtained by wearable sensors is closely related to sleep staging, thus allowing classification tasks to be accomplished.

In this work, we present the SleepGMUformer(Sleep-Gated Multimodal Units-former) sleep staging model. In terms of data, we concurrently harness both PSG and wearable device data. Since wearable sensors are easy to wear and use, they can fulfill the requirement of long-term sleep tracking in the future \cite{sleep}. This model performs well in analyzing sleep time data from wearable sensors, promoting sleep recognition in low-resource environments \cite{ref24}.

In terms of models, to enhance the ability to identify sleep stages and the generalization capability of the model, preprocessing methods that are more in line with real medical knowledge are needed to improve data quality and enhance features related to sleep staging \cite{ref25}. In this paper, de-trending of low-frequency trends that would dominate the entire EEG spectrum, removing low-frequency trends unrelated to brain activity, is used to reduce the impact of artifacts on EEG analysis \cite{ref26}. As well as temporal data processing of sleep recorded by wearable devices according to AASM rules. Additionally, the PSG channel signals corresponding to different sleep staging indicate different information and modalities. For example, EEG provides direct information about the state of brain activity and is particularly important in distinguishing between the diverse stages of NREM and REM phases, while EOG is particularly useful in identifying REM and wakefulness phases, which are both accompanied by rapid eye movements \cite{ref27}. We use the Gated Multimodal Units (GMU) module to improve the interpretability and accuracy of the sleep staging model and to allow a better understanding of the relevance of each modality of each instance.As shown in  Figure 1., through the integration of SleepGMUformer, we can achieve the full coverage of sleep staging test in municipalities and medically backward areas, as well as tracking and evaluation of the effect of the later diagnosis and treatment. Meanwhile, because SleepGMUformer can achieve more accurate sleep staging results, sleep staging can be applied to medicine, personal health management, business and other aspects in the future. 

SleepGMUformer is an attention-based dynamic fusion temporal network. In the preprocessing module, de-trending is performed on the EEG channel data of PSG, and alignment and missing value processing are carried out on the time series data recorded by wearable devices that are more in accordance with sleep division rules and the clinical medical requirements. For PSG data, there are two common input forms: one is to input one-dimensional raw signals, and the other is to output time-frequency images. The method we used is more in line with the requirements of clinical medicine. For sleep PSG data, there are two commonly used forms of input, one is to input one-dimensional raw signals and the other is to input time-frequency images. The time-frequency image is obtained by time-frequency analysis of the original signal, such as short-time Fourier transform (STFT), Choi-Williams distribution (CWD), and continuous wavelet transform (CWT) \cite{ref28}. 

However, our approach commences with the de-trending of the EEG signal derived from the raw PSG data, followed by a comprehensive analysis of the raw PSG signal to generate the time-frequency representation, which will be utilized as the input dataset. The time-series data, captured by the wearable device and segmented into 30-second epochs in accordance with the established sleep stage segmentation protocols, are synchronized temporally and refined to address any gaps in the data. The refined raw data is then projected onto a consolidated dimensional space before being introduced to a temporal feature extraction module that leverages a self-attention mechanism to identify salient temporal features. Upon the extraction of each feature modality, an advanced post-dynamic fusion strategy is deployed to integrate the modalities and derive more nuanced representations. For the synthesis of the ultimate representation that amalgamates all modalities, Gated Multimodal Units (GMU) equipped with highly interpretable gating mechanisms ascertain the significance of each modality concerning the sleep staging outcome. Empirical evidence garnered from two renowned public datasets, namely SleepEDF-78 and WristHR-Motion-Sleep, corroborates that SleepGMUformer surpasses current state-of-the-art methodologies. A synopsis of our principal contributions is delineated below:

1) Wearable sensors record time series data containing sleep information such as heart rate and respiration, and are introduced into sleep staging tasks, breaking through the current situation where sleep staging models only focus on PSG data and have poor generalization ability, and achieving compatibility in processing low dimensional and low volume data.

2) The GMU module is proposed to perform post-fusion, effectively allowing the model to weight modalities at the instance level and gain a deeper understanding of the correlation between each modality in each instance, thereby solving the problem of different contributions of each modality to identifying specific sleep stages. Provide better interpretable mechanisms while improving their performance.

3) De-trending processing for EEG use and alignment and missing value processing for wearable device sleep data, which are targeted preprocessing methods proposed for various sleep data more relevant to medical knowledge. According to the 1/f characteristic of the EEG spectrum, power decreases with increasing frequency, and the slow components that are rarely used for analysis in the signal are removed, emphasizing the fast components. The resulting prominent power change in the spectrum is the closest match to the original reality of the signal change actually observed by sleep experts when analyzing the original signal \cite{ref26}. For sleep data recorded by wearable devices, heart rate, breathing, and step count feature information are aligned in the time dimension according to AASM rules, and missing values are processed using interpolation methods.

\section{METHOD}
\label{sec1}
\subsection{Model overview}
\label{subsec1}

\begin{figure}[H]
\centering
\includegraphics[width=\columnwidth]{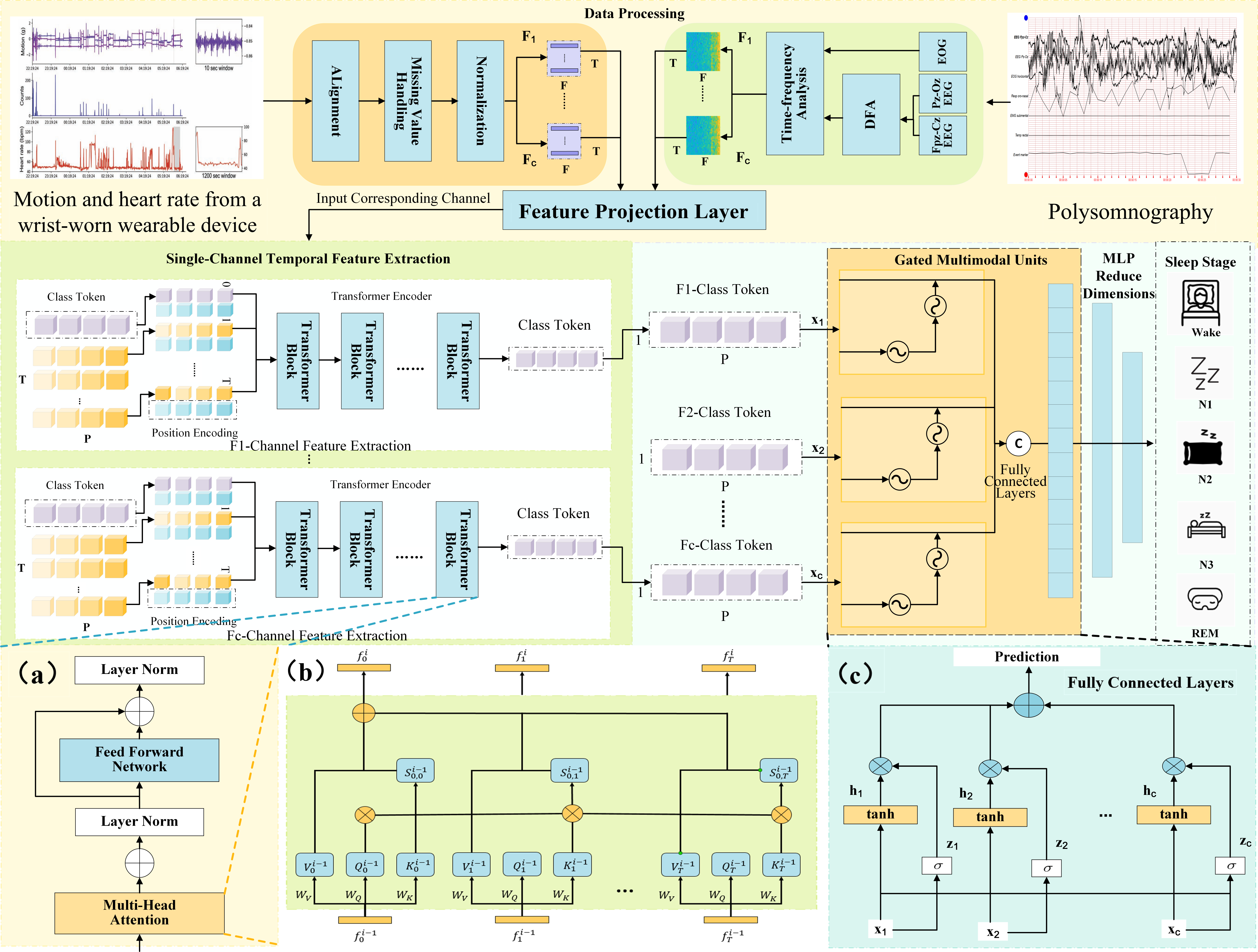}
\caption{SleepGMUfomer structure. Initially, the raw data are preprocessed by de-trending or alignment, missing value processing, etc. before being input to the single-channel feature extraction module, (a) Transformer Block module for feature extraction, where (b) Multi-head Attention module captures features from different angles by different "heads". Then, the features of each channel are input to the (c) Gated Multimodal Units (GMU) for dynamic multi-channel feature fusion. Finally, MLP is used to complete the classification.}
\label{fig1}
\end{figure}

Firstly, given a dataset $\{S_{n}\}_{n=1}^{N}$ of size N, $S_{n}=(\{X_1^{(n)},\ldots,X_{C}^{(n)}\},y^{(n)}))$ is the $n^{th}$ in a 30-second period containing the C channel data sets. In the SleepEDF-78 data set, $X_i^{(n)}\in\mathbb{R}^{T\times F},1\leqslant i\leqslant C$, denotes a time-frequency image extracted from a 30-second period EEG or EOG that can effectively represent specific wave and frequency components, where T is a time frame value of 29 and F is a frequency interval value of 128. In the WristHR-Motion-Sleep dataset, $X_{i}^{(n)} \in \mathbb{R}^{T \times F}, \quad 1 \leqslant i \leqslant C$, represents the sleep data in terms of inner heart rate or respiration data or number of steps per 30s, $X_{i} = \left\{ x_{j} \in \mathbb{R}^{N} \mid j = 1, 2, \ldots, T \right\}$, with T denoting the subsequence length, $x_{j} = \left\{ \alpha_{1}, \alpha_{2}, \ldots, \alpha_{F} \right\}$, and F is the number of signals collected per epoch. In addition, $y^{(n)} \in \{0,1\}^K$, y is the label, where K = 5 represents the five stages of sleep staging. 

Figure 2. shows the model architecture of SleepGMUformer, which consists of 1) single-channel temporal feature extraction; 2) multi-channel dynamic feature fusion; and 3) classification. For each modal channel, a temporal feature extraction module based on a self-attention mechanism extracts feature mappings from time-frequency images or timing data. These are independent so that they can adaptively capture different features from multimodal physiological signals. In the Multi-Channel Dynamic Feature Fusion module, post-fusion is performed by adding a GMU module, which effectively allows the model to weight modalities at the instance level, improving its performance while providing a better interpretability mechanism. Two fully connected layers and a softmax activation function at the end of the network complete the classification.

\begin{figure}[H]
    \centering

    \includegraphics[width=0.9\columnwidth]{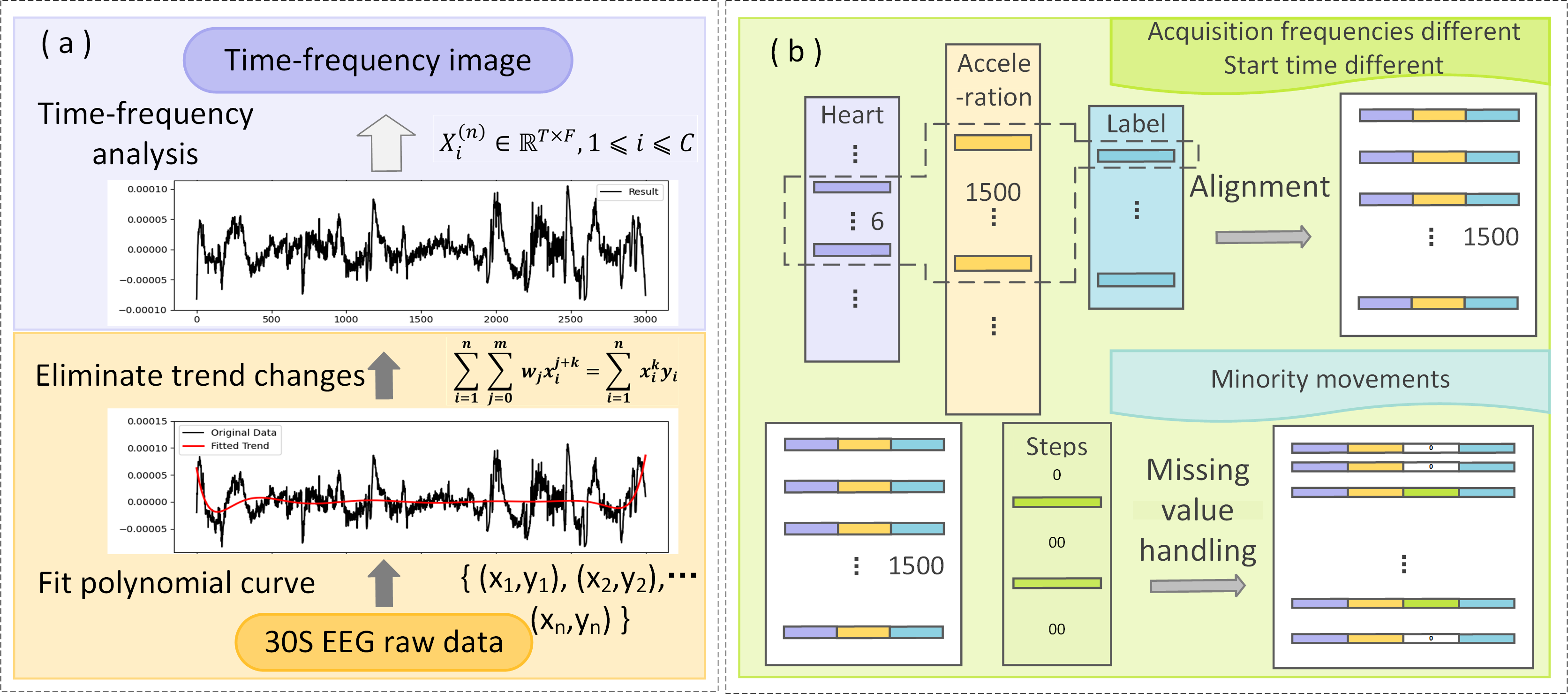}
    \caption{Preprocessing procedure on the datasets SleepEDF-78 and WristHR-Motion-Sleep. Preprocessing is performed on each 30-second unit of data. (a): de-trending process for EEG, (b): The wearable device records time-series data within the same 30 seconds, i.e., within the same sleep label, for alignment and missing value processing.}
    \label{fig:enter-label}
\end{figure}
\subsection{Preprocessing}
\label{subsec1}
For the SleepEDF-78 dataset, firstly, sampled records labeled as “motion” and “unknown” in the original dataset were removed prior to the experiment. Then, according to the convention established in previous studies, 30-minute sleep phases (60 minutes in total) were removed from the recordings when subjects got out of bed, as these sleep phases were labeled as wakefulness. Subsequently, N3 and N4 were uniformly labeled as N3 to obtain the categorical label $y \in \{W, N1, N2, N3, REM\}$, following the guidelines provided in the AASM manual. Finally, the EEG data were de-trended.

The principle of Detrended Fluctuation Analysis (DFA) is to characterize the data by subtracting a fitted straight line from the dot plot image of the original data to eliminate the trend changes presented by the data. The polynomial curve is fitted using the least squares method to obtain the fitted curve of the trend of the image, and the principle of polynomial fitting based on the least squares method is deduced as follows:

Firstly, $P = \{(x_1, y_1), (x_2, y_2),\ldots, (x_n, y_n)\}$, $P$ is a set of points that we assume, where the functional equation for \(x, y\) satisfies: \(f(x_i) = y_i\), \(i = 1, 2, \ldots, n\). Suppose the \(m\)-order polynomial function is:

\begin{equation}
\begin{aligned}
f(x_{i}, w_{j}) &= w_{0} + w_{1}x_{i} + w_{2}x_{i}^{2} + \cdots + w_{m}x_{i}^{m} \\
&= \sum_{j=0}^{m} w_{j} x_{i}^{j}
\end{aligned}
\end{equation}

where w is the polynomial coefficient. If this m-order polynomial is to be used to represent the relationship between x and y, then the error between the polynomial value and the true value is 
\begin{equation}
\begin{aligned}
e_{i} = f\left(x_{i}\right) - f\left(x_{i}, w_{j}\right) = y_{i} - \sum_{j=0}^{m} w_{j} x_{i}^{j}
\end{aligned}
\end{equation}

Polynomial fitting is performed using the least squares method to find the optimal set of polynomial coefficients to minimize the total error of the entire set of points after fitting. The problem of minimizing the total error can be transformed into minimizing the sum of squares of the errors. To obtain the sum of squares of the error for the entire set of points, the error sum and minimum are obtained by taking the partial derivative of $W_{k}$ and making it zero
\begin{equation}
\begin{aligned}
E(w)=\sum_{i=1}^{n}(\sum_{j=0}^{m}(w_{j}x_{i}^{j}-y_{i}))^{2_{\psi}}
\end{aligned}
\end{equation}
\begin{equation}
\begin{aligned}
\frac{\partial E(w)}{w_k}=2\sum_{i=1}^{n}\sum_{j=0}^{m}((w_{j}x_{i}^{j}-y_{i})x_{i}^{k})=0_{}
\end{aligned}
\end{equation}

The derivation of the formula gives:
\begin{equation}
\begin{aligned}
\sum_{i=1}^{n}\sum_{j=0}^{m}w_{j}x_{i}^{j+k}=\sum_{i=1}^{n}x_{i}^{k}y_{i}k=0,1,2,...,m_{}
\end{aligned}
\end{equation}

Writing it in matrix form gives:

\begin{equation}
\scriptsize
\begin{bmatrix}
n & \sum_{i=1}^n x_i & \sum_{i=1}^n x_i^2 & \dots & \sum_{i=1}^n x_i^m \\
\sum_{i=1}^n x_i & \sum_{i=1}^n x_i^2 & \sum_{i=1}^n x_i^3 & \dots & \sum_{i=1}^n x_i^{m+1} \\
\sum_{i=1}^n x_i^2 & \sum_{i=1}^n x_i^3 & \sum_{i=1}^n x_i^4 & \dots & \sum_{i=1}^n x_i^{m+2} \\
\vdots & \vdots & \vdots & \vdots & \vdots \\
\sum_{i=1}^n x_i^m & \sum_{i=1}^n x_i^{m+1} & \sum_{i=1}^n x_i^{m+2} & \dots & \sum_{i=1}^n x_i^{2m}
\end{bmatrix}
\begin{bmatrix}
w_0 \\
w_1 \\
w_2 \\
\vdots \\
w_m
\end{bmatrix}
=
\begin{bmatrix}
\sum_{i=1}^n y_i \\
\sum_{i=1}^n x_i y_i \\
\sum_{i=1}^n x_i^2 y_i \\
\vdots \\
\sum_{i=1}^n x_i^m y_i
\end{bmatrix}
\end{equation}

Calculated through math $\sum_{i=1}^{n} x_{i}^{j}, \quad j=0,1,2,\ldots, 2m$ and $\sum_{i=1}^{n} x_{i}^{j} y_{i} = 0, \quad j = 0, 1, 2, \ldots, m$, substituting into the above equation, the polynomial coefficients w are obtained and the fitted curve is obtained.

The image after detrending is obtained by subtracting the value of the vertical coordinate under the corresponding horizontal coordinate of the resulting fitted curve from the value of the vertical coordinate of the corresponding point on the dot plot of the original data. Then, after the original data were detrended, according to the AASM scoring manual, EEG, EOG, EMG, and major body movements are the basis for sleep staging, and specific wave and frequency components are important features. Such as low-frequency components in the 4-7 Hz range are frequently observed in stage N1, and sleep spindle (SS) or K-complex (KC) waves are hallmarks of stage N2. Finally, by applying STFT and logarithmic scaling, we convert the raw signals from each channel into a time-frequency image as an input to the model, which can effectively represent those specific wave and frequency components, and the whole process is shown in Figure 3(a).

For the WristHR-Motion-Sleep dataset: heart rate data is acquired every 5s, respiration data is acquired every 0.02s, and labeling follows the AASM sleep stage division rule every 30s. Since the data acquisition frequency of each modality of sleep is different, and the start time of each modality acquisition is also different, it is necessary to process the raw data with feature alignment and missing values. Firstly, We filtered out 1500 respiratory data and 6 heart rate data contained within the same 30s in which each label was located according to the label division time. Then, missing values are filled for the label at the corresponding moment of each respiratory data within the 30s while missing values are filled at the corresponding moment for the heart rate data within the same 5s in which the corresponding moment of each respiratory data is located. Finally, a few movements are recorded during sleep, and the corresponding moments of the supplemental tracts are recorded as 0 if there are none, as shown in Figure 3(b).

Considering the uncertainty of the measured magnitude and the possibility of outliers, the raw data need to be preprocessed to fit a normal distribution by normalizing the values of its elements into the [0,1] partition to increase the stability of the prediction. The preprocessing process is expressed as
\begin{equation}
\left\{
\begin{aligned}
& \tilde{\chi} = \mathcal{S}\big(\mathcal{N}(\mathcal{X})\big) \\
& \mathcal{N}(A) = \frac{A - A_{\text{min}}}{A_{\text{max}} - A_{\text{min}}} \\
& \mathcal{S}(A) = \frac{A - \mu}{\sigma}, \quad A = \left[a_{i,j}\right]_{n \times m}
\end{aligned}
\right.
\end{equation}

in Eq.(7) $\mathcal{N}_,\mathcal{S}$ describes the computation process of the normalization and standardization methods, respectively, A represents a two-dimensional matrix, and $\mu$ and $\sigma$ are the mean and variance of all the elements in matrix A. The data $\tilde{\mathcal{X}}=[\tilde{x}_{1},\tilde{x}_{2},\ldots,\tilde{x}_{T}]^{\mathrm{T}}$ is obtained, and the shape of $\mathrm{T}\times\mathrm{F}$ is consistent with that of the original data.

Finally, before the data from the two datasets are sent to the next module for single-channel temporal feature extraction, they need to pass through a feature mapping layer consisting of Multi-Layer Perception (MLP), which maps the low-dimensional raw data into a uniform high-dimensional space of dimension P. This is the first step in the process. $\tilde{F}=\tilde{\chi}\cdot w, \tilde{F}\in\mathbb{R}^{T\times P}$ high-dimensional feature passing matrix with weighting matrix ${\mathcal W}\in\mathbb{R}^{F\times P}$ matrix multiplication is performed to realize the dimensional transformation, in which the parameters of each moment are shared with each other and do not change in the time direction.
\subsection{Single-channel temporal feature extraction}
Based on the computational principle of signal frequency attention based mechanism, the single-channel temporal feature extraction module, which is centered on the attention mechanism, has the ability to mine temporal global dependency information from the input data in parallel. With the single-channel temporal feature extraction module, features relevant to sleep stage classification are extracted from a two-dimensional feature tensor ${\tilde{F}}\in\mathbb{R}^{T\times P}$. The extractor consists of multiple Transformer blocks stacked together, and each Transformer block as shown in Figure 2(a), assumes the function of extracting features layer by layer.

As shown in Figure 2, to satisfy the input specifications of the signal frequency attention based mechanism module, firstly, we adjust the input to divide the tokens along the time dimension and add positional information describing the temporal relationships between the tokens. Attention mechanism to discover the dependencies between tokens by similarity calculation.As shown in equation (8),$\tilde{F}=[\tilde{f}_{1},\tilde{f}_{2},...,\tilde{f}_{T}]^{\mathrm{T}}\in\mathbb{R}^{T\times P}$ is divided into T tokens along the principal dimensions, each of which represents a time-step feature vector. There is an additional learnable token $\tilde{f}_{0}$ as a CLASS Token spliced with $\tilde{F}$ for the final classification task. Since the transformer encoder eschews recursion and adds position encoding to the original input to introduce temporal information, the position encoding tensor $\mathrm{PE}\in\mathbb{R}^{(T+1)\times P}$ is computed, and $\mathrm{PE}$ is added to the T+1 feature tokens by an element-by-element addition operation to obtain a token $F=\{f_{0},f_{1},f_{2},...,f_{T}\}$, and the resulting F is input into a Transformer encoder consisting of multiple Transformer blocks.
\begin{equation}
\left\{
\begin{aligned}
& F = PE + [\tilde{f}_0 \mid \tilde{F}]^{\mathrm{T}} \\
& PE_{(pos,2i)} = \sin\left(\frac{pos}{10000\overline{P}}\right) \\
& PE_{(pos,2i+1)} = \cos\left(\frac{pos}{10000\overline{P}}\right)
\end{aligned}
\right.
\end{equation}
$\mathrm{PE}$ denotes the position encoding matrix. We use sine and cosine functions to encode the position information where $\mathrm{pos}$ is the position, $0\leq pos\leq T+1$, $2i$ or $2i+1$ is the number of dimensions of the input and $0\leq2i\leq2i+1\leq p$. Each dimension of the position encoding corresponds to a sinusoidal waveform, which allows the model to easily learn the relative positions based on the features of the sine and cosine functions.

When a feature consisting of T+1 tokens $\left\{f_{0}^{i-1},f_{1}^{i-1},\ldots,f_{T}^{i-1}\right\}$ into the attention module of the i-th head, according to the formula (9), the calculation is carried out to get $\{f_{0}^{i},f_{1}^{i},...,f_{T}^{i}\}$ and then input into the attention module of the next head, the structure of the attention mechanism module Figure 2(b) is shown.
\begin{equation}
\left\{
\begin{aligned}
& f_{j}^{i} = \sum_{k=0}^{T} V_{j}^{i-1} S_{j,k}^{i-1} \\
& S_{j,k}^{i-1} = \text{softmax}\left(\frac{Q_{j}^{i-1} K_{j}^{i-1}}{\sqrt{d}}\right) \\
& V_{j}^{i-1} = \xi(f_{j}^{i-1} W_{V}) \\
& Q_{j}^{i-1} = \xi(f_{j}^{i-1} W_{Q}) \\
& K_{j}^{i-1} = \xi(f_{j}^{i-1} W_{K})
\end{aligned}
\right.
\end{equation}
where $W_V,W_Q,W_K\in\mathbb{P}^P$ , are the weight metrics corresponding to $f_{j}^{i-1}$, $\xi$ is the activation function, and d is the dimension of the feature for which the scaling operation is performed. As shown in the overall model Figure 2, after the last transformer block is processed, the entire temporal feature extractor output of CLASS Token $F_t\in\mathbb{R}^P$ indexed as 0 serves as the final discriminative feature for each channel.

\subsection{Multi-channel dynamic feature fusion}
In most of the previous sleep staging models, a fully connected layer is usually used to directly concatenate the features extracted from each channel, which we believe is not optimal because different modal channels correlate differently with different sleep stages. For sleep polysomnography data, EEG is the main indicator in non rapid eye movement sleep (NREM) EEG is dominated by theta waves without spindle or K complex waves during N1, EEG is dominated by spindle and K complex waves with less than 20\% delta waves during N2, and delta waves are dominated during N3. EOG, on the other hand, is the main indicator to distinguish between rapid eye movements and non-rapid eye movements. For the WristHR-Motion-Sleep dataset, non-rapid eye movement sleep (NREM) results in decreased heart rate, frequent postural adjustments, and uniform breathing. In the rapid eye movement sleep (REM) stage, breathing is slightly faster and irregular, and body temperature and heart rate are elevated, with relatively sedentary physical activity. and the fully connected layer cannot correctly weight the relevance of each modality. In this work, we propose to adjust by changing the connectivity fusion with the GMU module.

 The structure of the GMU module is shown in Figure 2(c), the GMU module receives the feature vector $x_i \in \mathbb{R}^P$ of modality i. Firstly, it computes the hidden features $h_i$, and then it computes the gates $z_i$ that control the contribution of each modality to the overall output of the GMU module. Lastly, it compares the gates $z_1,z_2,...,z_N$ with hidden features $h_1,h_2,...,h_N$ are fused, and since with $W_i \in \mathbb{R}^{\text{shared} \times p}$ all modalities have the same dimension, they can be directly weighted by the gates $z_i$. The resulting GMU module has a global view of all modalities and can realize dynamic fusion due to the different correlations of different modal channels with different sleep stages, with the following formula.
\begin{equation}
\left\{
\begin{aligned}
& h_i = \tanh(W_i x_i^T) \in \mathbb{R}^{shared} \\
& z_{i} = \sigma(W_{z_i} [x_{i}]_{i=1}^{N}) \in \mathbb{R}^{shared} \\
& h = \sum_{i=1}^{n} z_{i} \odot h_{i}
\end{aligned}
\right.
\end{equation}
where N is the number of modes, $[x_i]_{i=1}^N$ denotes the result of splicing from vector $x_1$ to $x_n$, and $\odot$ stands for matrix multiplication.
\subsection{Classifier}
The classifier consists of a block of two fully connected layers that feed the results of the multichannel dynamic feature fusion \cite{chen2023fuselgnet} into the classifier, in the first fully connected layer the ReLU activation function is used, followed by the dropout layer. The two fully connected layers select and combine the features learned in the previous structure, and finally the softmax function is used to generate the output probabilities of mutually exclusive classes for sleep stage classification.

\section{EXPERIMENTS}
\subsection{Datasets}

Two publicly available datasets i.e. SleepEDF-78 \cite{ref29} and WristHR-Motion-Sleep were used in this study.We used the Fpz-Cz EEG, Pz-Oz EEG, and ROC-LOC EOG (horizontal) channels in the PSG of the SleepEDF-78 and heart rate, respiration, and activity patterns recorded by a wearable device in the WristHR-Motion-Sleep to complete the sleep stage classification task.

SleepEDF-78: This is a subset of the Sleep - EDF extended dataset (2018 version) which has been extended to include 78 subjects between the ages of 25 and 101 with 153 overnight PSG sleep recordings. Similar to the SleepEDF-20, the researchers performed two nights of PSG recordings for each subject. Due to an equipment error, one recording was lost for each of the 13-, 36-, and 52-year-old subjects. Trained technicians manually performed sleep staging of the corresponding PSG signals according to R\&K criteria and categorized each 30-s sleep period into labels {W, N1, N2, N3, N4, REM, MOVEMENT, UNKNOWN}. 

WristHR-Motion-Sleep: subjects spent the night in the laboratory with an eight-hour sleep opportunity, during which the subjects' respiration and heart rate were recorded using a WristHR-Motion-Sleep while polysomnography was performed. The dataset contained 39 subjects, excluding four patients with data transmission errors, three patients with sleep apnea, and one patient with REM sleep behavior disorder, divided into sleep stages (0-5, WAKE = 0, N1 = 1, N2 = 2, N3 = 3, REM = 5).

\subsection{Parameters}
To extract the time-frequency image, the 30-second calendar element of the PSG signal is transformed into a time-frequency image by a 256-point STFT. A 2s Hamming window with 50\% overlap was chosen as the window function. We used logarithmic scaling on the spectrum to convert to a log-power spectrum. This produces an image $X\in\mathbb{R}^{T\times F}$ where F = 128 frequency bins and T = 29 time points. The time-frequency images computed from each channel were normalized to the zero mean and unit variance of all time-frequency pairs and input to SleepGMUformer.

In the transformer encoder of our network, each sublayer of the transformer encoder uses a dropout rate of 0.4, and we use H = 8 attention heads and 128 hidden units in the feedforward layer. For each channel, features are extracted using 3 transformer encoders. In the last two fully connected layers, 64 hidden units are used with a dropout rate of 0.5. 

The experiments were conducted on two databases, WristHR-Motion-Sleep and SleepEDF-78, respectively. The dataset is divided into three parts based on the experimental ratio:60\% as a training set, 20\% as a validation set, and 20\% for final testing. The division process is performed based on truth labels to ensure a balanced data distribution. An Adam optimizer with a learning rate of $5\times10^{-3}$ is used to train the network. The optimizer performs degree descent based on this loss function. Finally, the batch size used for training is 256.
\subsection{Performance assessment metrics}
We used accuracy, Kappa coefficient, Macro F1 score (MF1), average sensitivity, and average specificity as overall performance metrics, whereas category-specific performance was evaluated using the class-wise MF1s for values used in the evaluation of this experiment.

\subsection{Experimental results}
\begin{figure}[H] 
  \centering 
  \begin{minipage}{0.5\textwidth} 
    \includegraphics[width=\linewidth]{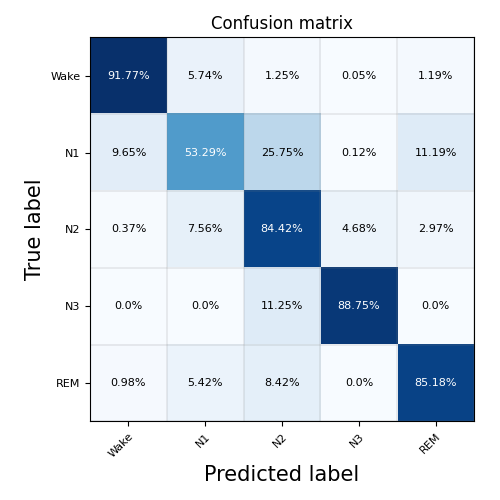} 
  \end{minipage}%
  \hfill 
  \begin{minipage}{0.5\textwidth} 
    \includegraphics[width=\linewidth]{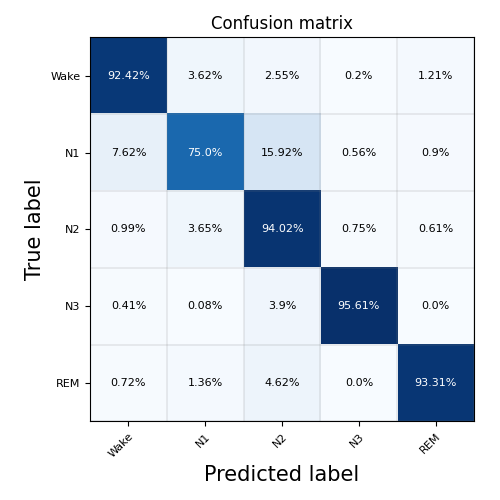} 
  \end{minipage}
  \caption{Confusion matrix on the dataset SleepEDF-78 and WristHR-Motion-Sleep, with the confusion matrix of SleepEDF-78 on the left and that of WristHR-Motion-Sleep on the right.} 
\end{figure}
\begin{table*}[ht]
\caption{Comparison of SleepGMUformer with results from SleepEDF-78 and other baseline models on the WristHR-Motion-Sleep dataset.}\label{tab1}
\centering
\scriptsize 
\setlength{\tabcolsep}{2pt} 
\renewcommand{\arraystretch}{1.1} 
\begin{tabularx}{\textwidth}{|p{1.5cm}|p{2.3cm}|*{10}{X|}}

\hline
\cline{1-12} 
\multirow{2}{*}{\textbf{Database}} & \multirow{2}{*}{\textbf{Model/System}} & \multicolumn{5}{c|}{\textbf{Overall metrics}} & \multicolumn{5}{c|}{\textbf{Class-wise MF1}} \\
\cline{3-12}
 &  & \textbf{Acc} & \textbf{K} & \textbf{MF1} & \textbf{Sens.} & \textbf{Spec.} & \textbf{Wake} & \textbf{N1} & \textbf{N2} & \textbf{N3} & \textbf{REM} \\
 \cline{1-12} 
\hline
\multirow{10}{*}{\makecell[c]{SleepEDF-\\78}} & SleepGMUformer & \textbf{85.0\%} & \textbf{0.9} & \textbf{79.2\%} & \textbf{78.8\%} & \textbf{96.2\%} & 91.8\% & \textbf{57.5\%} & 87.0\% & 80.8\% & 82.0\% \\
 & \makecell[l]{SleepTransfor-\\mer\cite{ref28}} & 81.4\% & 0.7 & 74.3\% & 73.4\% & 95.0\% & 91.7\% & 40.4\% & 84.3\% & 79.7\% & 77.2\% \\
 & XSleepNet2\cite{ref30} & 84.0\% & 0.8 & 77.9\% & 77.5\% & 95.7\% & 92.4\% & 47.0\% & \textbf{88.0\%} & 78.8\% & 81.3\% \\
 & XSleepNet1 & 83.6\% & 0.8 & 77.8\% & 77.2\% & 95.7\% & 92.0\% & 50.2\% & 85.9\% & 79.3\% & 81.3\% \\
 & SeqSleepNet\cite{ref14} & 82.9\% & 0.8 & 76.9\% & 75.9\% & 95.8\% & 89.3\% & 48.6\% & 85.6\% & 79.9\% & 79.3\% \\
 & FCNN+RNN & 82.9\% & 0.8 & 76.6\% & 75.9\% & 95.4\% & 92.5\% & 47.4\% & 85.6\% & 79.2\% & 80.0\% \\
 & U-Sleep\cite{ref31} & 81.9\% & - & - & - & - & \textbf{93.1\%} & - & 87.0\% & - & \textbf{88.0\%} \\
 & CNN-LSTM\cite{ref32} & 81.4\% & 0.7 & 73.5\% & - & - & \textbf{93.1\%} & 57.1\% & 71.0\% & - & - \\
 & AttnSleep\cite{ref33} & - & - & - & - & - & 92.0\% & 42.0\% & 85.1\% & \textbf{82.2\%} & 74.1\% \\
 & SleepEEGNet\cite{ref34} & 80.8\% & 0.8 & 77.3\% & - & - & 92.9\% & 49.1\% & 84.9\% & 80.7\% & 79.9\% \\
\cline{1-12} 
\multirow{8}{*}{\makecell[c]{WristHR-\\Motion-\\Sleep}} & SleepGMUformer & \textbf{94.5\%} & \textbf{0.9} & \textbf{89.6\%} & \textbf{90.0\%} & \textbf{97.9\%} & \textbf{92.0\%} & \textbf{70.0\%} & \textbf{94.2\%} & \textbf{96.5\%} & \textbf{95.6\%} \\
 & \makecell[l]{MLLR\cite{ref35}} & 90.7\% & 0.4 & - & 41.1\% & 94.6\% & - & - & - & - & - \\
 & \makecell[l]{BK-NN\cite{ref36}} & 90.9\% & 0.4 & - & 45.8\% & 95.0\% & - & - & - & - & - \\
 & \makecell[l]{B-RF\cite{ref37}} & 91.8\% & 0.4 & - & 51.4\% & 94.7\% & - & - & - & - & - \\
 & Neural net & 93.5\% & - & - & 52.3\% & 95.1\% & - & - & - & - & - \\
 & \makecell[l]{\makecell[l]{Epoch Cross-\\Transformer\cite{ref38}}} & 67.8\% & 0.4 & 56.3\% & 60.2\% & 96.6\% & 61.1\% & 28.6\% & 88.6\% & 60.4\% & 62.4\% \\
 & \makecell[l]{MultiChannel-\\SleepNet\cite{ref39}} & 83.5\% & 0.8 & 73.1\% & 71.6\% & 94.6\% & 77.6\% & 25.9\% & 86.7\% & 89.8\% & 85.1\% \\
 \cline{1-12}
\hline
\end{tabularx}

\end{table*}
As shown in Figure 4, the confusion matrix is used to show the performance of SleeoGMUformer on the datasets SleepEDF-78 and WristHR-Motion-Sleep, respectively, and we can see that the model SleeoGMUformer has excellent performance on both datasets and especially on the WristHR-Motion-Sleep dataset has the outstanding performance. We observe that most of the stages are accurately classified except for N1, possibly due to the fact that N1 is the first transition stage between wakefulness and sleep, and it has some of the same features as stages W and N2, which could be the reason for the lower performance of classifying N1 than the other sleep stages.
\subsubsection{Sleep staging performance}
{Table.1 shows the performance of SleeoGMUformer on the experimental dataset and compares it with previous works. Accuracy, Cohen's kappa (k), macro f1 score (MF1), average sensitivity and average specificity were used as overall performance metrics, while category-specific performance was assessed using the class-wise MF1s.

On the SleepEDF-78 database, our model has an overall accuracy of 85.03\% and a k value of 0.83. On the one hand, our model has an absolute improvement of 2.10\% in accuracy and 0.07 in k compared to SeqSleepNet.This result indicates that the methodology used by our model is more than SeqSleepNet has an advantage. On the other hand, the performance of our model is comparable to the existing state-of-the-art model XSleepNet2. In terms of accuracy, our model is 1.00\% higher than XSleepNet2, and in terms of k-value, it is 0.05 higher, showing that our model is competitive with XSleepNet2 in general. The class-wise MF1s further reveal the differences between them. Our model seems to favor sleep stages that are non-rapid eye movement stages (N1, N2, N3). In contrast, XSleepNet2 is superior in wakefulness stages (WAKE) and is comparable to and slightly superior to us in REM. Specifically, our model has an MF1 score of 57.45\% in the N1 phase, 87.04\% in the N2 phase, and 80.83\% in the N3 phase. and 91.84\% in the Wake phase. On the contrary, in the case of XSleepNet2, its MF1 score is 92.42\% in the Wake phase, which indicates that XSleepNet2 performs better in this phase.}

\subsubsection{Hypnogram}
\begin{figure}[H]
    \centering
    \includegraphics[width=0.9\textwidth]{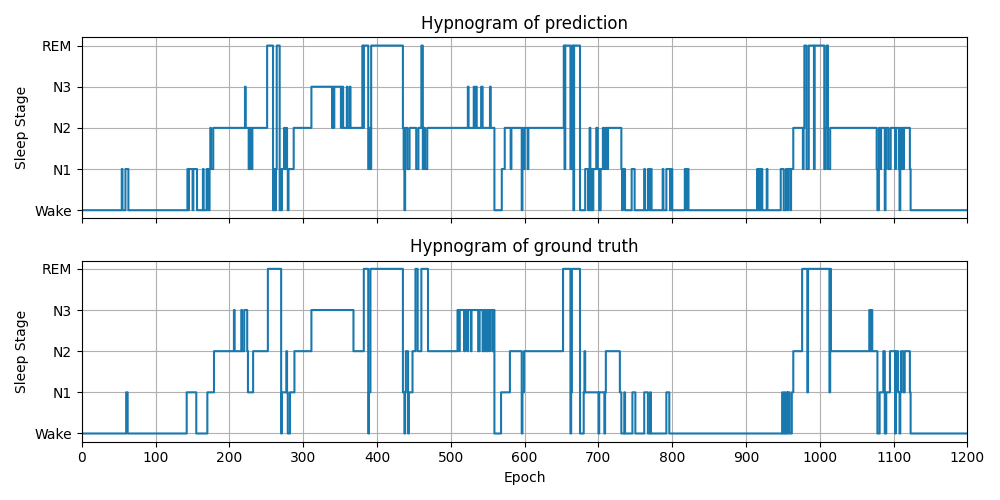} 
    \caption{Hypnogram showing classification accuracy; data from SleepEDF-78 dataset for subject A. Top row: classification results for sleep staging by the model, bottom row: ground truth of sleep stages}
    \label{fig:enter-label}
\end{figure}
{In the appendix Figure 5. shows a one-night hypnogram for subject A in SleepEDF-78, with ground truth in the lower panel and the predicted results of SleeoGMUformer in the upper panel. It can be seen that the predicted results are very similar to the ground truth and most of the misclassified sleep periods are N1, which is consistent with the experimental results in the table above. At the same time, we can see that the hypnogram produced by SleeoGMUformer may not be as smooth as the ground truth, and that classification anomalies can occur during sleep stage transitions because each sleep stage is shuffled and input to SleeoGMUformer without considering the temporal relationship between the epochs of a particular subject. Despite this limitation of such a design, it greatly reduces the complexity of the model while improving sleep staging performance.}
\subsubsection{Confidence estimation}
\begin{figure}[H]
    \centering
    \includegraphics[width=\textwidth]{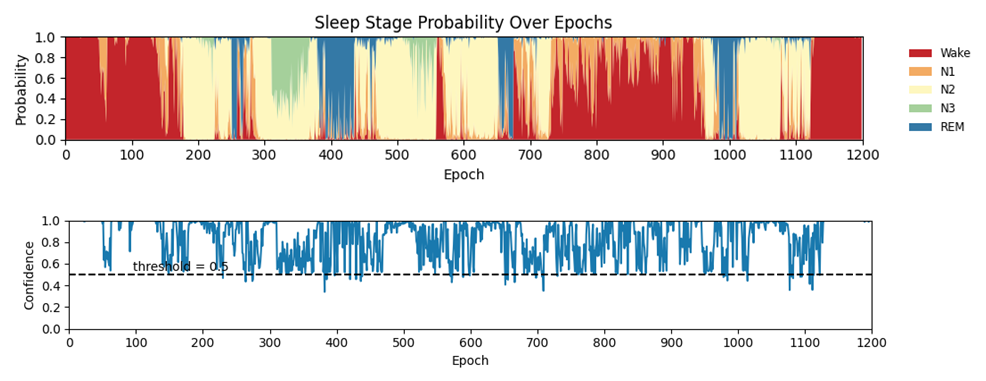} 
    \caption{Visualization of the estimated confidence for SleepEDF-78 subject A. Top row: quantitative confidence; bottom row: probability outputs}
    \label{fig:enter-label}
\end{figure}
{By analyzing the model's confidence in its predictions, it can help us better understand the model's performance \cite{ref40}. If the model has high confidence in most of the correct predictions and low confidence in incorrect predictions, this may indicate good model performance. In sleep staging, if the model has high confidence in the staging, then the physician or automated system can rely on this result with more confidence. Conversely, if the confidence level is low, then additional manual review or further testing is required \cite{ref41}. In previous studies, it has been found that times of misclassification are often associated with low confidence. For sleep in particular, classification of transitional epochs (whose sleep stages are different from their preceding and/or subsequent neighboring sleep stages) is often difficult \cite{ref42}.

We further demonstrate the above findings in Figure 6. In the figure, we perform a confidence quantification depicted along with the multi-class probability output for the above figure by performing a hypnogram to analyze a full night's sleep of subject A. The confidence quantification is depicted in the figure. With a confidence threshold of 0.5, we can see that the vast majority of the confidence levels are higher than the threshold, indicating that our model performs very well. Typically, on the other hand, low confidence tends to occur in the transition period between sleep stages, where the model has increased predictive uncertainty and also, by extension, the potential to be misclassified.
\begin{figure}[H]
\centerline{\includegraphics[width=0.9\columnwidth]{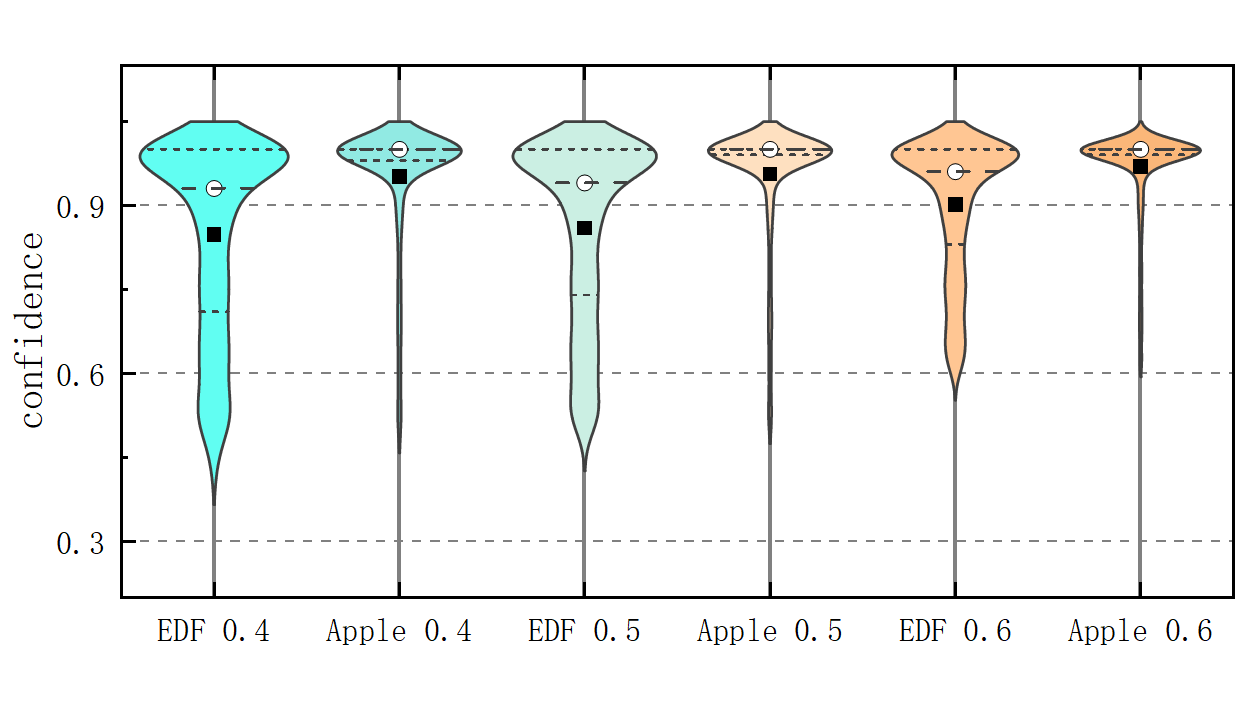}}
\caption{ Distribution of one-night confidence levels above confidence thresholds of 0.4, 0.5, and 0.6 for one subject in the SleepEDF-78 and WristHR-Motion-Sleep datasets.}
\label{fig3}
\end{figure}
}
Finally, we performed violin plot analysis on the confidence level data of subject A in the SleepEDF-78 dataset and subject B in the WristHR-Motion-Sleep dataset for a full night's sleep at the threshold of 0.4, 0.5 and 0.6, and above the threshold, respectively, which allows us to visualize the shapes of data distributions with different levels of confidence, and to compare the variability of the data distributions between them. As shown in Figure 7. it can be seen that the confidence level is mostly concentrated above 0.9, indicating that SleepGMUformer has a high level of confidence in the majority of correct predictions, suggesting that the model performs well and that physicians or automated systems can rely on this result with greater confidence. The figure represents the mean and the median, upon observation, it can be seen that in the WristHR-Motion-Sleep dataset, both the median and the average are higher than those in the SleepEDF-78 dataset, and the gap between the median and the average are smaller than those in the SleepEDF-78 dataset, which indicates that SleeoGMUformer has a higher level of confidence and a higher degree of concentration in dealing with the WristHR-Motion-Sleep dataset. This is also consistent with the experimental results that SleepGMUformer performs better in sleep stage classification on the WristHR-Motion-Sleep dataset than on the SleepEDF-78 dataset.
\subsection{Ablation experiment}
\begin{table}[H]
\caption{Performance comparison using different input channels on the SleepEDF-78 dataset}
\label{tab:sleep_stage_classification}
\centering

\renewcommand{\arraystretch}{1.0} 
\begin{tabularx}{\textwidth}{|p{2.5cm}|*{8}{>{\small\raggedright\arraybackslash}X|}}
\hline
\multirow{2}{*}{Input channels} & \multicolumn{3}{l|}{Overall metrics} & \multicolumn{5}{l|}{Per-class F1-score} \\ \cline{2-9} 
                                & Acc        & K          & MF1      & W   & N1   & N2   & N3  & REM \\ 
\cline{1-9}

\makecell[l]{EEG Fpz-Cz\\+EEG Pz-Oz}            & 76.2\%      & 0.7        & 65.5\%      & 92.4\% & 24.0\% & 76.5\% & 66.2\% & 68.2\% \\ \hline
\makecell[l]{EEG Fpz-Cz\\+EOG}                   & 83.5\%      & \textbf{0.8}        & \textbf{78.1\%}      & 93.4\% & 47.5\% & \textbf{85.6\%} & \textbf{82.6\%} & \textbf{81.6\%} \\ \hline
\makecell[l]{EEG Pz-Oz\\+EOG}                   & 83.0\%      & \textbf{0.8}        & 76.1\%      & \textbf{94.1\%} & 47.9\% & 84.7\% & 74.3\% & 79.3\% \\ \hline
Concatenation                   & \textbf{83.6\%}      & \textbf{0.8}       & 78.0\%      & 92.4\% &\textbf{51.8\%} & 83.6\% & 81.7\% & 80.9\% \\ \hline
\end{tabularx}

\end{table}
\begin{figure}[H]

\centerline{\includegraphics[width=0.9\columnwidth]{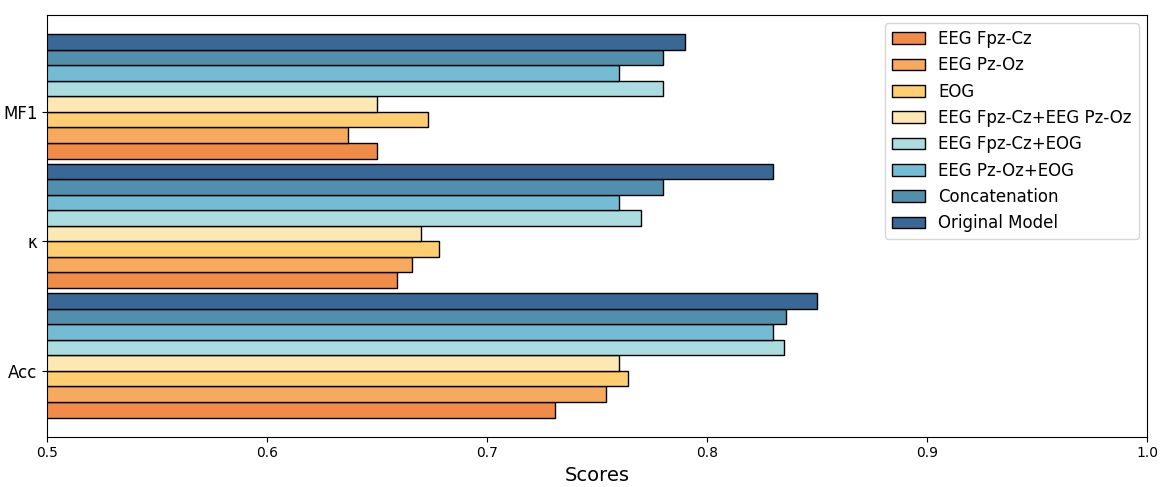}}
\caption{Performance comparison of the original model and variants using the SleepEDF-78 dataset}
\label{fig3}
\end{figure}

In order to prove the validity of our SleepGMUformer model, the module, we performed ablation experiments. We compared the original model in SleepEDF-78 as well as some variants on WristHR-Motion-Sleep, where other parameters are the same as the original model, respectively. These variants in SleepEDF-78 are described as follows.

1) EEG Fpz-Cz+EEG Pz-Oz: two-channel feature extraction block for processing only the Fpz-Cz EEG and Pz-Oz EEG channels.

2) EEG Fpz-Cz+EOG: two-channel feature extraction block, only for processing Fpz-Cz EEG and EOG channels.

3) EEG Pz-Oz+EOG: two-channel feature extraction block, only used to process Fpz-Cz EEG and EOG channels.

4) Concatenation: three channels are concatenated before input, no multi-channel feature fusion block.

5) Original Model: SleepGMUformer.

As Figure 8. shows, the original SleepGMUformer outperforms all these variants on the SleepEDF-78 dataset. Based on 1), 2), and 3), comparing our original model with those variants using dual channels, we can see from the analysis that the scheme of EEG Fpz-Cz + EEG Pz-Oz, which only takes into account the changes in EEG activity, performs poorly. In contrast, the addition of EOG resulted in a significant improvement in performance, based on studies showing that the contribution of EOG to improving the performance of the N1 phase and the REM phase is outstanding. This is in line with our results, as rapid eye movements are dependent on EOG monitoring. It also illustrates the importance of EOG in providing another dimension of information \cite{ref43}. The overall performance of the EEG Fpz-Cz + EOG scheme was slightly better than that of the EEG Pz-Oz + EOG in the conditions where EOG in considered. The overall performance of the EEG Fpz-Cz + EOG scheme was slightly better than that of the EEG Pz-Oz + EOG in the conditions where both EOGs were considered. According to the F1 scores for each level in Table 2, the Fpz-Cz showed a significant increase in the classification performance of the N2 and N3 phases, which may be due to the fact that the EEG Fpz-Cz channel typically covers the prefrontal region, a region that is associated with a wide range of cognitive functions, whereas the EEG Pz-Oz channel covers the central region and occipital lobe. The EEG Fpz-Cz channel, due to its location advantage, may be able to provide more useful information about sleep stage changes during feature extraction. For example, characteristic waveforms such as sleep spindle wave and K-complex wave may be more obvious in the Fpz-Cz channel, which in turn is one of the distinctive features of stages N2 and N3, which would be more helpful for the model to more accurately identify sleep stages. Another possibility is that the Fpz-Cz channel is potentially more complementary to the EEG Pz-Oz compared to the EOG channel, which requires further study in the future.

The above results show that the model with the multichannel feature fusion block outperforms the module without any single-channel variant, so we can conclude that fusing multichannel features by appropriate methods is effective for sleep staging. Another conclusion is that according to 4) the multichannel dynamic fusion block is able to effectively fuse features from different channels and is a key component of SleepGMUformer, resulting in better performance.
\begin{table}[H]
\caption{Performance comparison using different input channels on the WristHR-Motion-Sleep dataset}
\label{table:distribution}
\centering
\renewcommand{\arraystretch}{1.2} 
\begin{tabularx}{\textwidth}{|p{2.5cm}|*{11}{>{\small}X|}}
\hline
\multirow{2}{*}{Input channels} & \multicolumn{3}{c|}{Overall metrics} & \multicolumn{5}{c|}{Per-class F1-score} \\ \cline{2-9} 
                                & Acc        & K          & MF1       & W    & N1   & N2   & N3  & REM \\ 
\cline{1-9}

Heart+Steps                     & 69.9\%      & 0.5        & 59.7\%      & 62.1\% & 30.4\% & 77.9\% & 67.0\% & 61.1\% \\ \hline
Heart+Motion                    & 89.2\%      & 0.8        & 82.2\%      & 83.9\% & 47.5\% & 91.6\% & 95.5\% & 92.6\% \\ \hline
Motion+Steps                    & 87.8\%      & 0.8        & 80.4\%      & 77.9\% & 48.9\% & 92.2\% & 94.4\% & 88.9\% \\ \hline
Concatenation                   & \textbf{92.2\%} & \textbf{0.9} & \textbf{88.0\%} & \textbf{88.4\%} & \textbf{65.6\%} & \textbf{93.9\%} & \textbf{96.4\%} & \textbf{95.5\%} \\ \hline
\end{tabularx}
\renewcommand{\arraystretch}{1} 

\end{table}
\begin{figure}[H]
\centerline{\includegraphics[width=0.9\columnwidth]{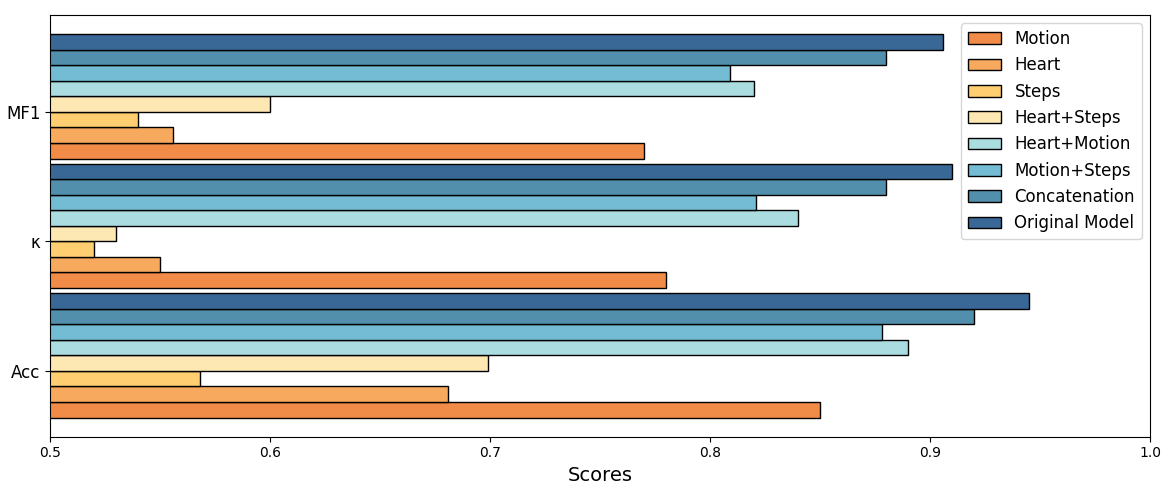}}
\caption{Performance comparison of the original model and variants using the WristHR-Motion-Sleep dataset}
\label{fig6}
\end{figure}
In WristHR-Motion-Sleep these variants are described as follows.

1) Heart + Steps: two-channel feature extraction block for processing only the Heart and Steps channels.

2) Heart + Motion: two-channel feature extraction block, only for processing only the Heart and Motion channels.

3) Motion + Steps: two-channel feature extraction block, only for processing only the Motion and Steps channels.

4) Concatenation: three channels are concatenated before input, no multi-channel feature fusion block.

5) Original Model: SleepGMUformer.

As the figure shows, the original SleepGMUformer outperforms all of these variants on the WristHR-Motion-Sleep dataset. Comparing our original model with those variants using dual channels according to 1), 2) and 3), the results show that one of the three physiological signals is indispensable, and the lack of any of the single-channel models results in a significant loss of performance, where the addition of the Motion channel, which is used to record respiration, significantly improves the results. Research has shown that breathing patterns are closely related to sleep stages. The frequency, depth and pattern of breathing change during different sleep stages. For example, breathing typically becomes slower and more regular during deep sleep, while breathing becomes more rapid and irregular during REM (rapid eye movement) sleep. Therefore, incorporating respiratory information into the model can provide more physiological signals that are directly related to sleep stages, thus improving the accuracy of sleep staging. Based on the F1 scores shown in Table 3, Heart+Motion outperforms Motion+Steps in conditions where both breathing is considered, and heart rate has significantly improved classification performance for W and REM stages. It was shown that heart rate variability (HRV) correlates well with the sleep stage. The characteristics of HRV vary in different sleep stages, especially in the W and REM stages. This is in line with our experimental results and makes our experimental results more interpretable. Finally, according to 4) at the same time the increase of multichannel dynamic fusion block can help the model to get better sleep staging performance.

\section{Discussion}
\begin{figure*} 
  \centering 
  \begin{minipage}{0.5\textwidth} 
    \includegraphics[width=\linewidth]{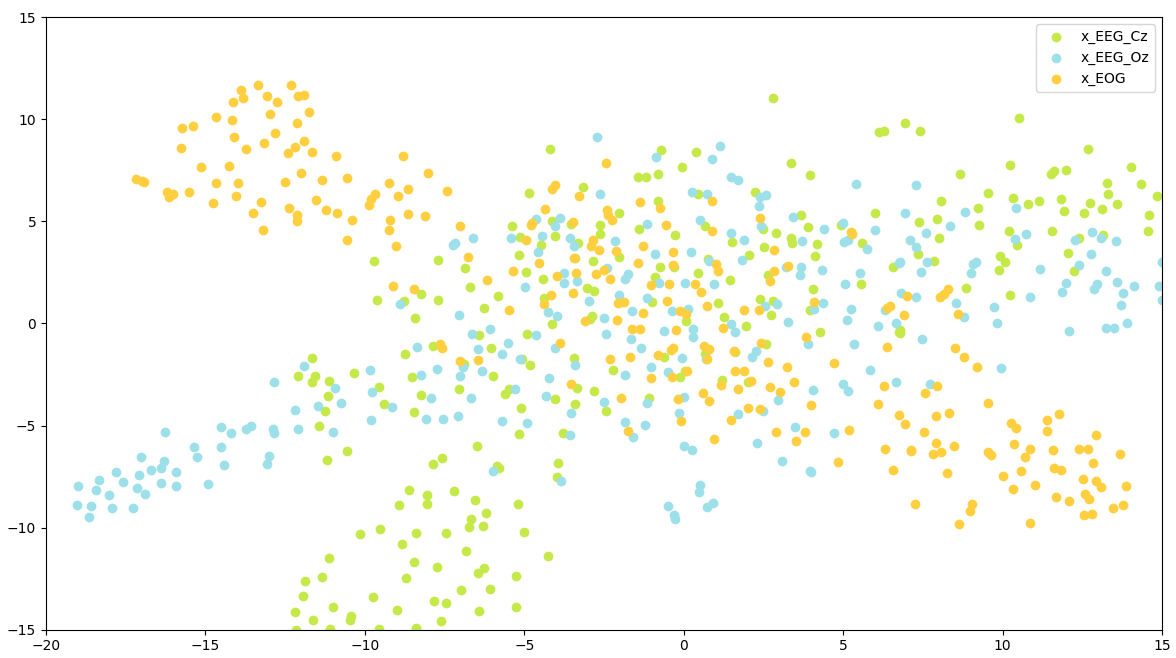} 
  \end{minipage}%
  \hfill 
  \begin{minipage}{0.5\textwidth} 
    \includegraphics[width=\linewidth]{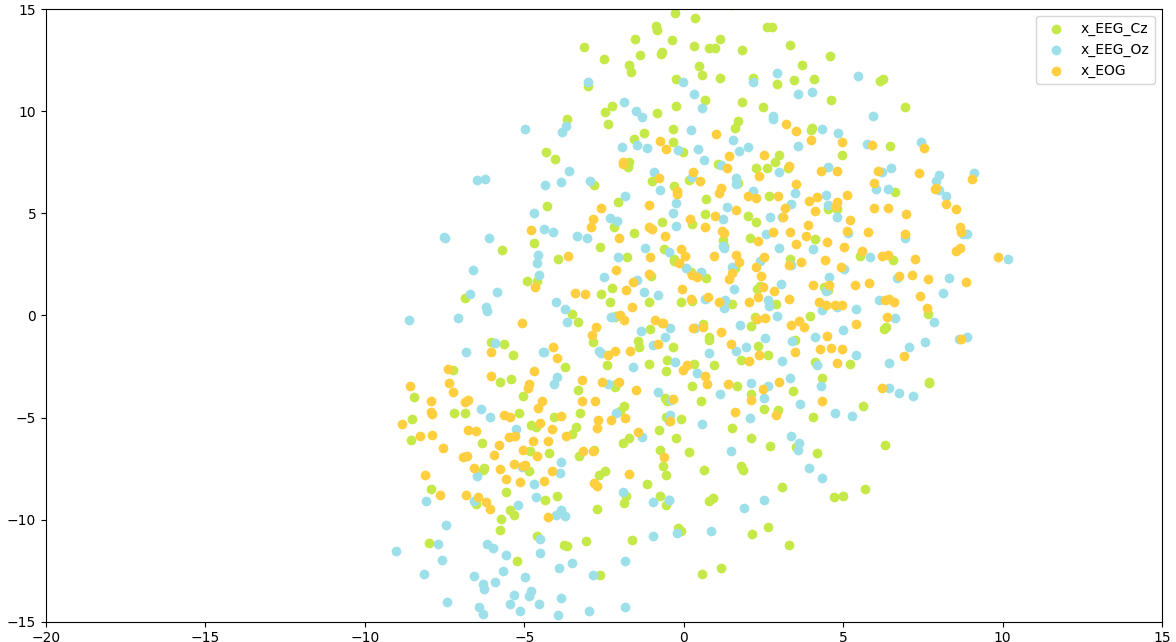} 
  \end{minipage}
  \caption{The t-SNE (t-distributed stochastic neighborhood embedding) plots of the EEG Fpz-Cz, EEG Pz-Oz, and EOG features in the SleepEDF-78 dataset.The image on the left is the t-SNE plots of raw input features, and the image on the right is the t-SNE plots of GMU module processed features} 
\end{figure*}
In our proposed SleepGMUformer model, initially, the STFT captures the frequency variations of the raw signals during different sleep stages and the spectral features containing the energy of each band associated with a particular sleep stage.  These are fed into a single-channel temporal feature extraction module, where multi-head attention assigns weights to each element of the different signal sequences, and features of different frequency components are learned by different heads, mining the complexity of the properties between different bands and undiscovered bands. The Layer Norm maintains the stability of the model due to the changes in the features in different sleep stages. The feed-forward layer performs nonlinear transformation and mapping of the features through layer-by-layer learning to capture the time-domain features (over-zero rate, standard deviation, etc.) and the frequency-domain features (power spectral density, etc.) of the signals in different sleep stages to fit more complex features, to enhance the model's ability to recognize the features in different sleep stages. Finally, the three features are fed into the GMU module for dynamic fusion of different modal channels with different correlations to different sleep stages.

To further identify the significant features that determine good classification, the extraction results of different features are visualized using t-SNE plots, where the more clustered the feature data points indicate that the model is more effective in extracting features, as shown in Figure 10. We present t-SNE plots for EEG Fpz-Cz, EEG Pz-Oz, and EOG features input as in Figure 10, t-SNE plots of features after feature extraction and after the completion of dynamic weight assignment in GMU as in Figure 10. It can be seen that the data points change from the initial dispersed distribution to a compact distribution, in which the EOG feature data points are the most clustered, which indicates that our SleepGMUformer model can extract the features of the sleep data efficiently and that the EOG contributes more significantly to the classification results. Combined with the results of the ablation experiment, Figure 8. shows that using only EOG for classification is not effective, which indicates that EOG contributes significantly to the classification results, but it still needs to be combined with other features to complete the classification.

Although our proposed model SleepGMUformer exhibits superior performance, some limitations need to be addressed. Firstly, SleepGMUformer exhibits a weak classification performance for stage N1 on both different datasets. The fact that N1 is the first transition stage between wakefulness and sleep leads to a lower classification performance for N1, but in clinical settings, stage N1 has obvious characteristics that distinguish it from other sleep stages, and the brain waves in stage N1 change compared to the wakefulness state (stage W), with the $\alpha$ -wave gradually weakening, and low-amplitude, mixed-frequency EEG activity replacing the $\alpha$ -wave in the wakefulness state, with a slowed down frequency of $\geqslant $1 Hz. The N1 stage usually does not contain the characteristic sleep spindle and K-complex waves of stage N2 or the slow waves ($\delta$ -waves) of stage N3. Therefore, theoretically, the classification performance of stage N1 can be completely improved, how to extract the features related to stage N1 for refinement and improve the classification performance of this stage will be our next step. Additionally, since our model only uses a single epoch as input without considering the context, therefore there are some sample-specific transition stage classification anomalies in the model output. Lastly, in this process, we utilized a rich dataset that included sleep data collected from wearable devices in addition to traditional sleep polysomnography data. But PSG data contains records of more channels, such as EMG and respiratory signals, temperature, etc. In the future, we will try to introduce these channel signals and study their impact on sleep stage classification performance.

\section{Conclusion}
In this study, we propose an innovative sleep staging model, SleepGMUformer, which is based on the signal frequency attention based mechanism and is designed to process multi-channel PSG data and heart rate and respiration data collected by wearable devices. Our model contains four main parts: a preprocessing module, a single-channel temporal feature extraction module, a multi-channel dynamic feature fusion module, and a classifier. 
The model breaks through the status quo of only processing PSG data, realizes the compatible processing of two kinds of heterogeneous data, and performs well on both datasets, showing high accuracy and K values reaching 85.0\% and 0.83 on the SleepEDF-78 dataset, and the overall accuracy and K values of the model on the WristHR-Motion-Sleep dataset even reached 94.5\% and 0.91, which indicates that the model can effectively handle physiological data from different sources. The high accuracy and kappa value indicate that the model not only shows excellent classification performance but also has high consistency with experts' classification results, which indicates that the sleep staging results provided by our model are accurate and can be trusted. 

However, the model also suffers from, lower classification accuracy in the N1 phase, not taking into account the temporal relationship between time points and the low number of channels in the PSG data used. In the future, we will deeply analyze the characteristics of the N1 stage and explore more effective ways to improve the model's classification performance for this stage. Attempts will be made to introduce temporal context information between epochs to improve the model's classification accuracy during sleep stage transitions. The inclusion of more channels of signals such as EMG, respiratory signals, and body temperature will be considered to enhance the model performance.

\section*{Acknowledgments}
This research was supported in part by National Natural Science Foundation of China under Grant No.62106117, Natural Science Foundation of Shandong Province under Grant No.ZR2021QF084.



\end{document}